\documentclass[10pt,twocolumn,letterpaper]{article}

\usepackage{iccv}
\usepackage{times}
\usepackage{epsfig}
\usepackage{graphicx}
\usepackage{subcaption}
\usepackage{amsmath}
\usepackage{amssymb}
\usepackage{algorithm}
\usepackage{algorithmic}
\usepackage{bbm}
\usepackage{amsfonts}
\usepackage{mathrsfs}
\usepackage{multirow}

\iccvfinalcopy 

\def\iccvPaperID{0} 

\begin{document}

\title{Identity-Aware U-Net: Fine-grained Cell Segmentation via Identity-Aware Representation Learning}

\author{
Rui Xiao\\
South China University of Technology\\
xray13714753857@gmail.com
}

\maketitle
\thispagestyle{empty}

\begin{abstract}
Precise segmentation of objects with highly similar shapes remains a challenging problem in dense prediction, especially in scenarios with ambiguous boundaries, overlapping instances, and weak inter-instance visual differences. While conventional segmentation models are effective at localizing object regions, they often lack the discriminative capacity required to reliably distinguish a target object from morphologically similar distractors. In this work, we study fine-grained object segmentation from an identity-aware perspective and propose Identity-Aware U-Net (IAU-Net), a unified framework that jointly models spatial localization and instance discrimination. Built upon a U-Net-style encoder-decoder architecture, our method augments the segmentation backbone with an auxiliary embedding branch that learns discriminative identity representations from high-level features, while the main branch predicts pixel-accurate masks. To enhance robustness in distinguishing objects with near-identical contours or textures, we further incorporate triplet-based metric learning, which pulls target-consistent embeddings together and separates them from hard negatives with similar morphology. This design enables the model to move beyond category-level segmentation and acquire a stronger capability for precise discrimination among visually similar objects. Experiments on benchmarks including cell segmentation demonstrate promising results, particularly in challenging cases involving similar contours, dense layouts, and ambiguous boundaries.\end{abstract}

\section{Introduction}

Precise object segmentation and localization is a fundamental problem in computer vision, with broad applications in biomedical image analysis, industrial inspection, remote sensing, autonomous systems, and visual surveillance. In many real-world scenarios, however, the challenge goes beyond separating foreground objects from the background. The model must further distinguish a target object from other visually similar instances, such as people with similar appearance, cells with nearly identical morphology, or objects with highly overlapping contours and textures. In such cases, accurate segmentation requires not only strong spatial localization, but also fine-grained discriminative capability.

Deep neural networks have substantially advanced dense prediction in recent years. Fully convolutional architectures have established effective paradigms for semantic segmentation~\cite{long2015fcn}, while U-Net~\cite{ronneberger2015unet}, DeepLab~\cite{chen2018deeplabv3plus}, and Mask R-CNN~\cite{he2017maskrcnn} have demonstrated strong performance in pixel-level prediction, multi-scale context modeling, and instance-aware mask estimation. More broadly, deep learning has become a dominant paradigm in medical image analysis~\cite{litjens2017survey}. More recently, foundation-style segmentation models such as Segment Anything have shown impressive transferability across diverse visual domains~\cite{kirillov2023sam}. In biomedical imaging, methods such as StarDist~\cite{schmidt2018stardist}, SplineDist~\cite{mandal2020splinedist}, Cellpose~\cite{stringer2021cellpose}, Hover-Net~\cite{graham2019hovernet}, and distance-map-based nuclei segmentation methods~\cite{naylor2019distance} have further improved the segmentation of cells and nuclei by incorporating object shape priors, contour-aware modeling, and stronger instance separation mechanisms. Despite these advances, most existing methods remain primarily optimized for category-level delineation and boundary quality, rather than explicit discrimination between highly similar target and non-target instances.

This limitation becomes particularly critical in fine-grained segmentation settings. In densely packed biomedical images, neighboring nuclei or cells often exhibit highly similar morphology and weak boundaries, which makes accurate instance separation especially challenging and has motivated contour-aware and shape-aware models such as DCAN~\cite{chen2016dcan}, StarDist~\cite{schmidt2018stardist}, and SplineDist~\cite{mandal2020splinedist}. More generally, distinguishing between morphologically similar cells in microscopy images, or separating a specific object from a set of shape-similar distractors, is often difficult when supervision is based solely on pixel-wise classification. Standard segmentation networks tend to emphasize shared category-level features, which can weaken their ability to preserve subtle instance-specific differences. As a result, the learned representation may be sufficient for identifying \emph{where an object is}, but insufficient for determining \emph{which object is the desired target}. This problem is further aggravated by noisy backgrounds, ambiguous boundaries, dense layouts, and the presence of hard negatives that resemble the target in contour or texture.

In parallel, metric learning and contrastive learning have demonstrated strong discriminative power in recognition and retrieval tasks. Siamese networks~\cite{bromley1993siamese} established an early framework for similarity learning, and FaceNet~\cite{schroff2015facenet} showed that triplet-based objectives can effectively enforce compact intra-target representations and separated inter-target embeddings. Subsequent studies have further highlighted the importance of hard negative mining for fine-grained discrimination~\cite{hermans2017triplet}, while supervised contrastive learning has shown that explicitly structuring the embedding space can substantially improve representation quality~\cite{khosla2020supcon}. These results suggest that explicitly modeling similarity and dissimilarity relations may also benefit dense prediction, especially when the primary challenge lies in distinguishing a target object from visually similar distractors.

Motivated by these observations, we revisit fine-grained object segmentation from an identity-aware perspective and propose Identity-Aware U-Net (IAU-Net), a unified framework that jointly models spatial localization and instance discrimination. Built upon a U-Net-style encoder--decoder architecture, our method augments the segmentation backbone with an auxiliary identity-aware embedding branch that learns discriminative target representations from high-level features. In addition to conventional segmentation supervision, we introduce contrastive metric learning with noisy backgrounds and morphology-similar negative samples, encouraging the network to refine target-specific features while suppressing confusing distractors. In this way, the segmentation branch provides precise spatial guidance, whereas the identity-aware branch improves the ability to distinguish the desired object from nearby or visually similar alternatives.

Compared with conventional segmentation models, IAU-Net does not treat dense prediction as a purely category-level problem. Instead, it reformulates precise segmentation as a joint task of localization and fine-grained discrimination. This makes the framework particularly suitable for benchmarks involving morphologically similar objects, including but not limited to cell segmentation. More broadly, the proposed formulation provides a simple and extensible direction for fine-grained object segmentation in scenarios where accurate target selection is as important as accurate boundary prediction.

Our main contributions are summarized as follows:
\begin{itemize}
    \item We revisit fine-grained object segmentation from an identity-aware perspective and highlight the limitation of conventional segmentation models in distinguishing morphology-similar targets and distractors.
    \item We propose IAU-Net, a unified architecture that combines a U-Net-style segmentation branch with an auxiliary identity-aware embedding branch for joint localization and discrimination.
    \item We introduce contrastive metric learning with noisy backgrounds and hard negative samples to refine target-specific features and improve robustness against visually similar objects.
    \item We show that the proposed framework is applicable not only to cell-related benchmarks but also to broader object segmentation settings involving strong inter-instance similarity.
\end{itemize}

\section{Methods}

We propose Identity-Aware U-Net (IAU-Net), a unified framework for fine-grained object segmentation in scenarios where the target object must be distinguished from visually or morphologically similar distractors. The key idea is to augment a standard U-Net segmentation backbone with an auxiliary identity-aware embedding branch, and to jointly optimize pixel-level segmentation and triplet-based metric learning. In this way, the model learns not only accurate object localization, but also fine-grained discrimination between the target and similar negatives.


\subsection{Overall Framework}

Our training pipeline is built on triplet supervision. Given an input image with pixel-level annotations, we construct training triplets as follows: the anchor image $I_a$ denotes the full input image that contains the target instance, together with its corresponding segmentation mask $M_a$; the positive reference $I_p$ is another crop sharing the same target identity; and the negative reference $I_n$ is sampled from a different identity, such as noisy backgrounds or adjacent objects with similar morphology. Depending on the dataset configuration, these negative samples can be selectively drawn from a dedicated hard-negative pool. This formulation explicitly drives the model to learn fine-grained, target-specific discrimination in conjunction with dense prediction.

\begin{figure*}[t]
    \centering
    \includegraphics[width=\linewidth]{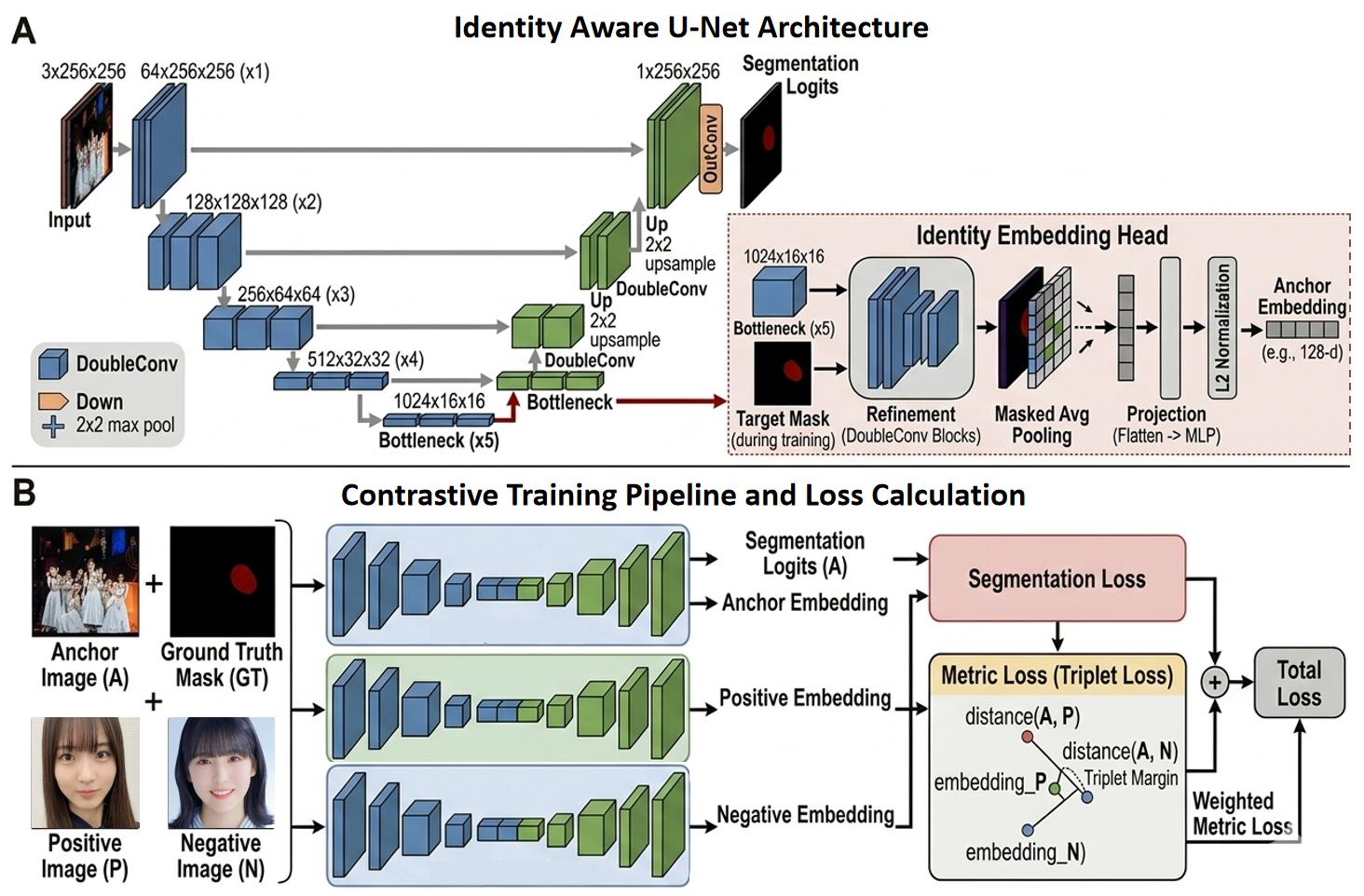}
    \caption{Overview of the proposed Identity Aware U-Net architecture. The model consists of a shared encoder and a segmentation decoder pathway, together with skip connections between corresponding stages. An auxiliary identity-aware embedding branch is attached to the bottleneck feature to learn discriminative target representations under triplet-based supervision.}
    \label{fig:architecture}
\end{figure*}

During training, the anchor image is forwarded through the full network to produce both a segmentation prediction and an identity embedding. In contrast, the positive and negative reference images are only used for identity learning and therefore pass through the shared encoder and embedding head without entering the segmentation decoder. The segmentation branch encourages precise object localization, while the metric branch enforces discriminative separation between similar objects in the learned feature space.The network architecture is illustrated in Figure~\ref{fig:architecture}.

\subsection{Identity-Aware U-Net Architecture}

Our model is built upon a U-Net-style encoder--decoder architecture with skip connections. The encoder consists of an initial double convolution block followed by four downsampling stages. Each stage contains two $3 \times 3$ convolution layers, each followed by batch normalization and a ReLU activation. Downsampling is performed with $2 \times 2$ max pooling. As the spatial resolution decreases along the encoder, the number of feature channels increases progressively, enabling the network to capture increasingly abstract semantic information.

The decoder follows the standard U-Net design. Starting from the bottleneck representation, the feature map is progressively upsampled back to the input resolution. Each upsampling stage is followed by feature fusion with the corresponding encoder feature map through skip connections, and then refined by a double convolution block. Depending on the implementation setting, upsampling is performed either by bilinear interpolation or transposed convolution. A final $1 \times 1$ convolution layer projects the decoder output to the desired number of segmentation classes.

The skip connections are essential to the network design. They transfer high-resolution spatial features from the encoder to the decoder, allowing the model to recover object boundaries and fine local details that may otherwise be lost during repeated downsampling. As a result, the segmentation branch can produce more precise masks, especially in cases with thin structures, ambiguous boundaries, or densely packed objects.

To enhance discriminative capability, we attach an auxiliary identity-aware embedding head to the bottleneck feature of the U-Net. This branch shares the encoder with the segmentation decoder but has a different objective. Instead of reconstructing a dense mask, it learns a compact embedding that represents the target object's identity in feature space. The embedding head first refines the bottleneck feature with additional convolutional layers. For the anchor image, the ground-truth target mask is resized to the bottleneck resolution and used to suppress irrelevant background responses, so that the embedding is computed primarily from the target region. For positive and negative reference images, where no segmentation mask is required during forward prediction, the embedding is obtained directly from the encoded features using global feature aggregation. The pooled representation is then passed through a lightweight projection multilayer perceptron and normalized to produce the final identity embedding.

This design allows the shared encoder to support two complementary tasks. The decoder learns spatially precise dense prediction, while the identity branch encourages the encoder to preserve subtle target-specific characteristics that are useful for distinguishing similar instances. In effect, the segmentation network is no longer optimized purely for category-level mask prediction, but also for instance-level discrimination.


\subsection{Triplet-based Metric Learning}

To explicitly enforce fine-grained discrimination, we train the embedding branch with triplet loss. For each triplet $(I_a, I_p, I_n)$, the network produces an anchor embedding $\mathbf{e}_a$, a positive embedding $\mathbf{e}_p$, and a negative embedding $\mathbf{e}_n$. The triplet objective encourages the anchor to be closer to the positive sample than to the negative sample by a margin $m$:
\begin{equation}
\mathcal{L}_{\mathrm{tri}}
=
\max \left(
\|\mathbf{e}_a - \mathbf{e}_p\|_2
-
\|\mathbf{e}_a - \mathbf{e}_n\|_2
+
m,\,
0
\right).
\end{equation}

This formulation is particularly suitable for our task because the negative examples are often visually similar to the target object. By forcing the embedding space to separate such hard negatives, the network becomes more robust to confusing instances that share similar contours, morphology, or texture patterns.

The segmentation branch is supervised using a combination of pixel-wise classification loss and Dice loss. For binary segmentation, the segmentation objective is defined as
\begin{equation}
\mathcal{L}_{\mathrm{seg}}
=
\mathcal{L}_{\mathrm{bce}}(\hat{Z}_a, M_a)
+
\mathcal{L}_{\mathrm{dice}}(\sigma(\hat{Z}_a), M_a),
\end{equation}
where $\hat{Z}_a$ denotes the predicted segmentation logits for the anchor image, $M_a$ is the ground-truth mask, and $\sigma(\cdot)$ is the sigmoid function. For multi-class segmentation, the binary classification term is replaced by cross-entropy:
\begin{equation}
\mathcal{L}_{\mathrm{seg}}
=
\mathcal{L}_{\mathrm{ce}}(\hat{Z}_a, M_a)
+
\mathcal{L}_{\mathrm{dice}}(\mathrm{softmax}(\hat{Z}_a), M_a).
\end{equation}

The full training objective is given by
\begin{equation}
\mathcal{L}_{\mathrm{total}}
=
\mathcal{L}_{\mathrm{seg}}
+
\lambda \mathcal{L}_{\mathrm{tri}},
\end{equation}
where $\lambda$ balances segmentation supervision and identity-aware metric learning.

Through this joint optimization, the segmentation branch improves spatial precision and mask quality, while the triplet loss encourages the shared encoder to preserve subtle target-specific cues. As a result, the network becomes better at distinguishing the desired object from noise-like backgrounds or morphology-similar distractors, leading to more accurate fine-grained segmentation and localization.

\subsection{Training and Inference}

During training, the anchor image is processed by both the segmentation decoder and the identity-aware embedding branch, whereas the positive and negative reference images are processed only by the embedding pathway. The model is trained end-to-end with online data augmentation, including random flipping, affine perturbation, and color jitter. Optimization is performed using RMSProp with gradient clipping, and mixed-precision training can be enabled for computational efficiency.

During inference, only the input image is required. The network predicts the segmentation mask through the segmentation branch, while the identity-aware supervision from training remains embedded in the shared encoder features. Although the embedding branch is primarily introduced for training, it improves the discriminative quality of the learned representation and thereby enhances segmentation performance in challenging scenarios involving highly similar objects.

\section{Experiments}

\subsection{Datasets}

We evaluate the proposed method on two public nucleus segmentation benchmarks, namely BBBC038v1~\cite{caicedo2019nucleus} and MoNuSeg~\cite{kumar2020multiorgan}. These two datasets were selected because they exhibit substantial variation in image appearance, object density, boundary ambiguity, and inter-instance similarity, making them suitable benchmarks for evaluating fine-grained segmentation and localization.

\textbf{BBBC038v1.}
BBBC038v1 is a large-scale nucleus segmentation benchmark derived from the 2018 Data Science Bowl and released through the Broad Bioimage Benchmark Collection~\cite{caicedo2019nucleus}. It contains 2D light microscopy images collected from more than 30 biological experiments under diverse imaging conditions, cell types, staining protocols, and acquisition settings. Such high diversity makes BBBC038v1 particularly challenging for segmentation models, as nuclei can vary significantly in size, shape, intensity, and crowding patterns across experiments.

\textbf{MoNuSeg.}
MoNuSeg is a multi-organ nucleus segmentation benchmark for digital pathology~\cite{kumar2020multiorgan}. The dataset consists of H\&E-stained histopathology image patches extracted from whole-slide images at $40\times$ magnification. The original challenge dataset contains 30 training images of size $1000 \times 1000$ with 21,623 manually annotated nuclei, collected from multiple organs including breast, liver, kidney, prostate, bladder, colon, and stomach. Compared with fluorescence microscopy data, MoNuSeg presents a different set of challenges, including larger appearance variation across tissue types, more complex nuclear morphology, staining inconsistency, and dense cellular distributions.

\textbf{Cross-validation protocol.}
To obtain a robust and fair evaluation, we perform 10-fold cross-validation on each dataset. Specifically, for each benchmark, the available annotated images are randomly partitioned into ten folds at the image level. In each run, nine folds are used for training and the remaining fold is used for testing. Final performance is reported by averaging the results over all ten folds. This evaluation protocol reduces the bias introduced by a particular train--test split and provides a more reliable assessment of the generalization ability of the proposed method across diverse imaging conditions and object morphologies.

\subsection{Comparison with U-Net}

To validate the effectiveness of the proposed identity-aware embedding branch, we compare our IAU-Net with a standard U-Net baseline on the BBBC038v1 benchmark. The two models share the same encoder--decoder backbone and segmentation objective, while the only architectural difference is that our method introduces an additional identity embedding head together with triplet-based metric supervision. This comparison is designed to isolate the contribution of the identity-aware branch and to verify whether it improves fine-grained object discrimination beyond conventional pixel-wise segmentation.

Table~\ref{tab:bbbc038_unet_vs_iaunet} reports the quantitative results on BBBC038v1. Compared with the vanilla U-Net, the proposed IAU-Net achieves higher Dice and IoU scores, demonstrating that the identity-aware embedding branch provides more discriminative feature representations for segmentation. By explicitly encouraging the network to separate target objects from morphology-similar negatives, the proposed method improves boundary localization and preserves finer local structures in crowded scenes.

\begin{table}[t]
\centering
\caption{Comparison between U-Net and our IAU-Net on BBBC038v1. The best results are shown in bold.}
\label{tab:bbbc038_unet_vs_iaunet}
\begin{tabular}{lcc}
\hline
Metric & U-Net & IAU-Net \\
\hline
Dice & 0.8262 & \textbf{0.8485} \\
IoU  & 0.7353 & \textbf{0.7571} \\
\hline
\end{tabular}
\end{table}

\begin{figure}[t]
    \centering
    \begin{subfigure}[b]{0.48\linewidth}
        \centering
        \includegraphics[width=\linewidth]{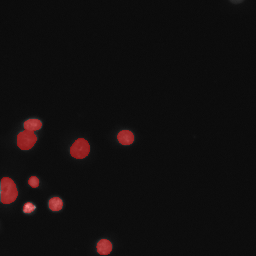}
        \caption{U-Net}
        \label{fig:overlay_unet}
    \end{subfigure}
    \hfill
    \begin{subfigure}[b]{0.48\linewidth}
        \centering
        \includegraphics[width=\linewidth]{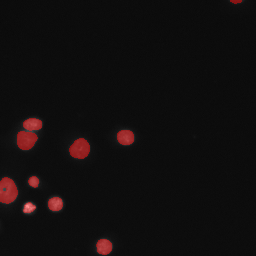}
        \caption{Identity-Aware U-Net}
        \label{fig:overlay_iaunet}
    \end{subfigure}
    \caption{Qualitative comparison on BBBC038v1. The left figure shows the prediction of the standard U-Net, while the right figure shows the prediction of the proposed Identity-Aware U-Net. The proposed method better separates touching and adherent cells under weak or ambiguous boundaries.}
    \label{fig:qualitative_unet_vs_iaunet}
\end{figure}

Beyond the quantitative gains, our method also demonstrates improved capability in separating touching and adherent cells compared with the standard U-Net. While specialized instance segmentation methods such as SplineDist~\cite{mandal2020splinedist} are specifically designed for this purpose and typically achieve stronger instance separation performance, our method still provides meaningful improvements without requiring instance-level annotations. This setting is also related to recent efforts on annotation-efficient cell segmentation, such as Scribble2Label~\cite{lee2020scribble2label}, which reduce the dependence on dense manual labels. As illustrated in Figure~\ref{fig:qualitative_unet_vs_iaunet}, the standard U-Net tends to merge adjacent cells into a single connected region when their boundaries are weak or highly ambiguous. In contrast, the proposed IAU-Net produces more discriminative predictions and shows better capability in separating adherent cells, although perfect instance separation remains difficult without explicit instance-level supervision.

Overall, these results indicate that the proposed identity-aware formulation is beneficial for improving global segmentation accuracy and can also provide improved discrimination between adjacent instances. However, we emphasize that our method is not designed as a dedicated instance segmentation approach, and complete instance separation remains a challenging problem that may require additional instance-level supervision.

\subsection{Comparison with other methods}

We further compare the proposed method with other weakly supervised segmentation methods on the MoNuSeg benchmark. In particular, we follow the comparison setting of Scribble2Label (S2L)~\cite{lee2020scribble2label} and report the results of representative scribble-supervised baselines, including GrabCut, Pseudo-Label, pCE Only, rLoss, and Scribble2Label. To enable a direct comparison under the same weak-supervision setting, we additionally include our method, IAU-Net, in the same benchmark protocol.

Table~\ref{tab:monuseg_comparison_scribble} summarizes the comparison results on MoNuSeg. This comparison helps position the proposed method against existing annotation-efficient segmentation approaches on a challenging nucleus segmentation benchmark.

\begin{table}[t]
\centering
\caption{Comparison with weakly supervised methods on MoNuSeg.}
\label{tab:monuseg_comparison_scribble}
\begin{tabular}{lcc}
\hline
Method & Dice & IoU \\
\hline
GrabCut~\cite{rother2004grabcut}            & 0.1534 & 0.0703 \\
Pseudo-Label~\cite{lee2013pseudolabel}      & 0.6113 & 0.5607 \\
pCE Only~\cite{tang2018regularized}         & 0.6319 & 0.5766 \\
rLoss~\cite{tang2018regularized}            & 0.6337 & 0.5789 \\
Scribble2Label~\cite{lee2020scribble2label} & 0.6408 & 0.5811 \\
IAU-Net (Ours)                              & \textbf{0.7299} & \textbf{0.5835} \\
\hline
\end{tabular}
\end{table}

Compared with existing scribble-supervised baselines, our method is designed to enhance fine-grained discrimination by introducing an identity-aware embedding branch and triplet-based metric supervision. This is particularly relevant for MoNuSeg, where nuclei often appear densely packed and exhibit substantial morphological similarity. By explicitly encouraging the network to separate target-consistent features from confusing negatives, the proposed IAU-Net provides improved robustness in challenging scenarios involving clustered nuclei and ambiguous boundaries.

\section{Conclusion}

In this paper, we presented Identity Aware U-Net, a fine-grained segmentation framework designed to improve the discrimination and localization of morphologically similar objects. Built upon a U-Net-style encoder--decoder architecture, the proposed method augments the segmentation backbone with an identity-aware embedding branch and introduces triplet-based metric supervision to enhance feature separability. In contrast to conventional segmentation models that primarily focus on category-level pixel prediction, our method explicitly encourages the network to distinguish target-consistent objects from confusing negatives with similar shape or appearance.

Extensive experiments on BBBC038v1 and MoNuSeg demonstrated that the proposed method achieves consistent improvements over a standard U-Net baseline. In particular, the quantitative and qualitative results showed that Identity Aware U-Net provides stronger boundary localization and better separation of touching or adherent nuclei. These findings suggest that incorporating identity-aware representation learning into dense prediction is an effective way to improve fine-grained segmentation performance, especially in crowded scenes with ambiguous boundaries and strong inter-instance similarity.

Another notable advantage of the proposed framework is that it improves instance discrimination without relying on manual instance-level annotations. This makes the method particularly relevant for biomedical image analysis, where dense annotation is expensive and time-consuming. By introducing an identity-aware objective into the segmentation process, our framework offers a simple yet effective alternative to more annotation-intensive pipelines.

In future work, we plan to extend this framework in several directions. First, more advanced metric learning objectives and hard-negative mining strategies may further strengthen feature discrimination. Second, query-conditioned or prompt-guided variants could improve the flexibility of the framework for interactive and target-specific segmentation. Finally, we believe that the proposed identity-aware formulation can be generalized beyond nucleus segmentation to broader fine-grained object segmentation tasks in which accurate target selection is as important as accurate boundary prediction.

{\small
\bibliographystyle{ieee_fullname}
\bibliography{main}
}

\end{document}